\documentclass[10pt,journal,compsoc]{IEEEtran}
\ifCLASSOPTIONcompsoc
  \usepackage[nocompress]{cite}
\else
  \usepackage{cite}
\fi
\ifCLASSINFOpdf
   \usepackage[pdftex]{graphicx}
   \graphicspath{{../pdf/}{../jpeg/}}
   \DeclareGraphicsExtensions{.pdf,.jpeg,.png}
\else
   \usepackage[dvips]{graphicx}
   \graphicspath{{../eps/}}
   \DeclareGraphicsExtensions{.eps}
\fi
\usepackage{amssymb}
\usepackage{amsmath}
\usepackage{algorithmic}
\usepackage{algorithm}

\usepackage{array}

\ifCLASSOPTIONcompsoc
  \usepackage[caption=false,font=footnotesize,labelfont=sf,textfont=sf]{subfig}
\else
  \usepackage[caption=false,font=footnotesize]{subfig}
\fi
\usepackage{fixltx2e}
\usepackage{xcolor}
\usepackage{float}
\usepackage{ragged2e}

\begin{document}
%
\title{Semi-Supervised Adversarial Monocular \\Depth Estimation}
%
%
%
%

\author{
    Rongrong~Ji,  \emph{Senior Member}, \emph{IEEE},  Ke~Li$^*$, Yan~Wang, \emph{Member}, \emph{IEEE}, Xiaoshuai~Sun, \emph{Member}, \emph{IEEE}, Feng~Guo, \emph{Member}, \emph{IEEE}, Xiaowei Guo, Yongjian Wu, Feiyue Huang, and Jiebo~Luo, \emph{Fellow}, \emph{IEEE} 
    \IEEEcompsocitemizethanks{\IEEEcompsocthanksitem Rongrong Ji, Feng Guo and Xiaoshuai Sun are with Fujian Key Laboratory of Sensing and Computing for Smart City, School of Information Science and Engineering, Xiamen University, 361005, China. Yan Wang is with Microsoft. Jiebo Luo is with Department of Computer Science, 
   	University of Rochester. Ke Li, Xiaowei Guo, Yongjian Wu and Feiyue Huang are with Tencent Youtu Lab.\protect\\
	E-mail: tristanli@tencent.com
}
\thanks{Manuscript received on Oct 11, 2018, and revised on June 27, 2019.}
}

%
%

\markboth{IEEE Transactions on Pattern Analysis and Machine Intelligence}%
{Shell \MakeLowercase{\textit{et al.}}: Bare Demo of IEEEtran.cls for Computer Society Journals}
%



\newcommand{\yan}[1]{\textcolor{blue}{{[Yan: #1]}}}
\newcommand{\ccom}[1]{\textcolor{red}{{[#1]}}}
\IEEEtitleabstractindextext{
\begin{abstract}
\justifying
In this paper, we address the problem of monocular depth estimation when only a limited number of training image-depth pairs are available. To achieve a high regression accuracy, the state-of-the-art estimation methods rely on CNNs trained with a large number of image-depth pairs, which are prohibitively costly or even infeasible to acquire. Aiming to break the curse of such expensive data collections, we propose a semi-supervised adversarial learning framework that only utilizes a small number of image-depth pairs in conjunction with a large number of easily-available monocular images to achieve high performance. In particular, we use one generator to regress the depth and two discriminators to evaluate the predicted depth , \emph{i.e.}, one inspects the image-depth pair while the other inspects the depth channel alone. These two discriminators provide their feedbacks to the generator as the loss to generate more realistic and accurate depth predictions. Experiments show that the proposed approach can (1) improve most state-of-the-art models on the NYUD v2 dataset by effectively leveraging additional unlabeled data sources; (2) reach state-of-the-art accuracy when the training set is small, \emph{e.g.}, on the Make3D dataset; (3) adapt well to an unseen new dataset (Make3D in our case) after training on an annotated dataset (KITTI in our case).
\end{abstract}

}

\maketitle

\IEEEdisplaynontitleabstractindextext

%
\IEEEpeerreviewmaketitle

\IEEEraisesectionheading{\section{Introduction}\label{sec:introduction}}

%
%
%
%

\IEEEPARstart{E}stimating scene depth is the foundation of various computer vision tasks.
Coming with the recent success in deep learning techniques \cite{eigen2014depth,laina2016deeper,hu2019revisiting}, vision-based depth estimation is a more flexible and affordable solution for depth acquisition comparing with using active sensors like LIDAR.
Among the vision-based approaches, monocular depth estimation has its unique advantages due to its extremely low requirement on the sensor, and thus has received extensive focus from both academy and industry.

However, the practical performance of monocular depth estimation retains unsatisfied in real-world applications. 
To achieve reasonable accuracy, early works relied on using visual cues such as shading \cite{prados2006shape} and texture \cite{aloimonos1988shape}, or using additional information to the image content such as camera motion \cite{hartley2003multiple} and multi-view stereo \cite{seitz2006comparison}.
However, such dependency on extra sensor information greatly weakens the unique advantages of monocular depth estimation.
Subsequently, learning based methods \cite{saxena2006learning,eigen2014depth,laina2016deeper} have been introduced and have attracted increasing research attention.
Among the earliest works, Saxena \textit{et al.} in \cite{saxena2006learning,saxena2009make3d} adopted superpixels together with Markov Random Fields (MRFs) and Conditional Random Fields (CRFs) to infer depth.
Non-parametric approaches, such as \cite{karsch2012depth}, aim to reconstruct the depth of a given image by warping the most similar images in a dataset and then transferring the corresponding depth map.
Notably, deep learning schemes have been introduced in depth estimation by Eigen \textit{et al.} \cite{eigen2014depth}, which have subsequently dominated the learning-based estimation methods.
In their settings, Convolutional Neural Networks (CNNs) are trained with meticulously designed loss functions to regress the depth value. However, such methods typically need a large (or at least sufficient) amount of image-depth pairs to effectively train the deep models. For instance, \cite{eigen2014depth}, \cite{fu2018deep} and \cite{hu2019revisiting} take more than 50K ground truth depth to sufficiently train their deep models.
Such a requirement is indeed problematic, \emph{i.e.}, the ground truth depth maps are typically expensive, and sometimes are even infeasible to acquire, due to the intrinsic limitations in depth acquisition approaches. 
For instance, most depth sensors are generally expensive with relatively low resolutions comparing to existing image sensors, and their depth sensing ranges are typically fixed and hard to adapt to different scenarios.
More recently, video sequences and stereo image pairs have been introduced as an alternative approach to generate depth pairs~\cite{zhou2017unsupervised,godard2017unsupervised}. However, these works instead introduce a new requirement to align videos or image pairs, which are therefore not suitable when only irrelevant images are available.

In this work, we propose a semi-supervised depth estimation framework to relax the need for large-scale supervised information, which instead only requires a small amount of image-depth pairs accompanied with a large amount of cheaply-available monocular images in training. Differing from the supervised training schemes in most existing models [1], [2], [12], such a semi-supervised setting imposes extra unsupervised cues.
In particular, to train the depth regressor, a generative adversarial learning paradigm is introduced, which contains a generator for depth estimation and two discriminators to evaluate the depth estimation quality and its consistency with the corresponding RGB image, respectively. The generator can be of any cutting-edge image-to-depth estimation models, \emph{e.g.}, \cite{laina2016deeper, hu2019revisiting, ronneberger2015u}.
One of the discriminators, called \emph{pair discriminator}, aims to distinguish images and their predicted depth maps from the real image-depth pairs.
The other discriminator, called \emph{depth discriminator}, aims to evaluate the quality of depth map, which tells whether the predicted depth is drawn from the same distribution of the real depth. During training, unannotated images are fed to the generator and then the network loss is computed according to the feedback from these two discriminators. The generator, the pair discriminator and the depth discriminator together simulate a Bayesian framework: The depth discriminator accounts for the \emph{priori} $p(d)$, and the pair discriminator accounts for the joint distribution $p(d, I)$. The \emph{likelihood} has $p(I|d) = p(d, I) / p(d)$. And the final posterior has $p(d|I) \propto p(I|d)p(d)$.

The contributions of our work are as follows:

\begin{itemize}
\item{We propose a semi-supervised framework to release modern models' reliance on image-depth pairs by leveraging unlabeled RGB images in the depth estimation task.}
\item{We implement the semi-supervised framework by an adversarial learning paradigm, in which a generator network estimates the depth while two discriminator networks inspect the estimated depth-image pair and depth, respectively.}
\item{The framework generalizes well to different network architecture. For example, the generator can be any cutting-edge depth estimator. Specifically, networks described in \cite{ronneberger2015u, laina2016deeper, hu2019revisiting} all receive a performance gain in our experiments. The semi-supervised framework can also be used in dataset adaptation.}
\end{itemize}

Thorough experimental validations are given in three folds:
\begin{itemize}
\item We demonstrated that the proposed framework is able to benefit most cutting-edge models \cite{laina2016deeper,hu2019revisiting,ronneberger2015u} when limited training data (image-depth pairs) are
available. On the NYUD dataset, the proposed framework is able to decrease
the evaluation errors of our baseline model \cite{hu2019revisiting} by $28\%$, $43\%$ and $38\%$ \textit{w.r.t}. the Rel, RMSE and $log_{10}$ error in a practical
setting that the labeled training is limited. We achieve this by leveraging $50\%$ more unlabeled images drawn from the same distribution and alternatively train the
generator with the supervised information (from L1 discrepancy) and unsupervised information (from discriminators’ genuineness feedback). 
\item We demonstrated that the proposed framework is able to reach state-of-the-art accuracy when the training set is small. On the Make3D dataset, our method reveals its potential on the small dataset and reach a state-of-the-
art evaluation error of Rel $0.158$, RMSE $6.137$ and $log_{10}$ $0.067$.
\item  We demonstrated that the proposed framework adapts well to unseen new datasets after training on an annotated one. When directly testing the model trained on KITTI on the Make3D dataset, our semi-supervised method overall performs the best with other supervised models.
\end{itemize}

The rest of our paper are organized as follows.
We introduce and discuss the related work in Sec. 2.
The proposed approach is then presented in Sec. 3.
We give details of the experimental settings and discussions in Sec. 4.
Finally, we conclude the paper in Sec. 5.  

\begin{figure}
	\centering\includegraphics[width=1\linewidth]{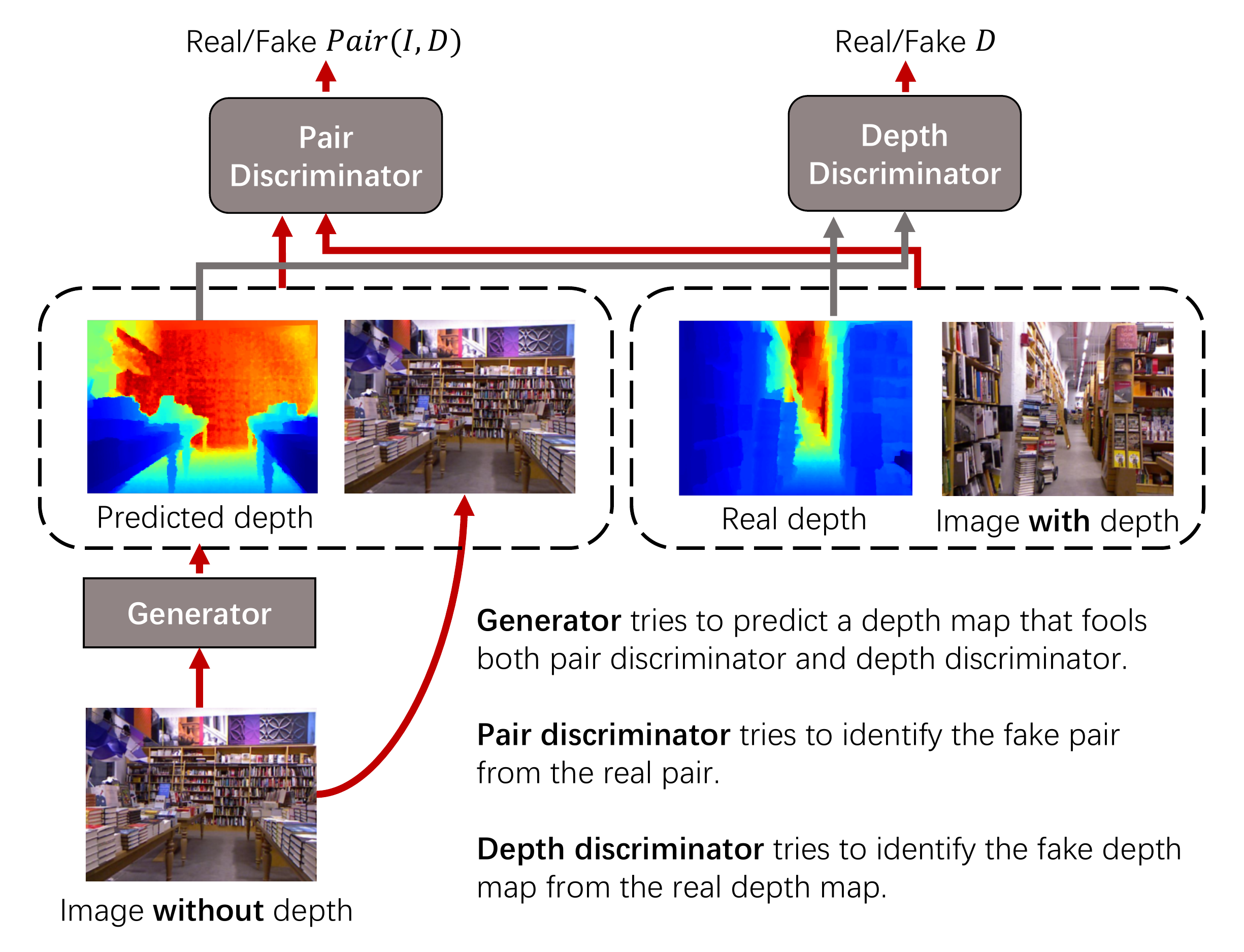}
	\caption{Our semi-supervised adversarial framework.
    We try to leverage a vast amount of unannotated images together with a small number of image-depth pairs to train a depth estimation network.
    The generator receives gradients from two discriminator networks to update its parameters.
    Unlike traditional losses such as $L_1$, $L_2$ and Huber norm, the loss from two discriminators' feedback can preserve better details with less requirement on the amount of ground truth.}
	\label{fig:pipeline}
\end{figure}

\section{Related Work}
\subsection{Geometry Depth Estimation}
Early works have revealed that 3D shape can be inferred from monocular camera images by shading \cite{prados2006shape}, texture \cite{aloimonos1988shape}, and motion \cite{hartley2003multiple}, \textit{etc}.  
For instance, the methods in \cite{prados2006shape,aloimonos1988shape,zhang1999shape} inferred 3D object shapes from shading or texture with the assumption that color and texture distribute uniformly.
The methods in \cite{nagai2002hmm,hassner2006example,han2003bayesian} studied the reconstruction of known object classes.
Levin \textit{et al.} \cite{levin2007image} used a modified camera aperture to predict depth by defocusing.
Structure from Motion (SfM) \cite{hartley2003multiple} and visual Simultaneous Localization And Mapping (vSLAM) \cite{klein2007parallel}  are in another direction, which have been popular algorithms in reconstructing 3D scene from multiple images.
SfM and vSLAM leverage camera motion to estimate camera poses first, and then infer depth by triangulating consecutive keyframe pairs.
The reconstructed points are accurate with respect to their relative positions, but only cover a very small portion of the entire scene, which leads to a sparse depth map that is hard to use in various applications.

\subsection{Learning-based Depth Estimation}
Among the earliest works in learning-based monocular depth estimation, Saxena \textit{et al.} \cite{saxena2006learning} predicted depth from local features, and then incorporated global context to refine the local prediction via an MRF.
This work was later extended in \cite{saxena2009make3d}, which reconstructs scene structure using the predicted depth map. 
Karsch \textit{et al.} \cite{karsch2012depth} used non-parametric feature matching to retrieve the nearest neighbors of the given images within an RGB-D dataset.
Then the corresponding depth maps of the retrieved images are warped and merged together to obtain the final visually pleasing depth estimation. 
Ladicky \textit{et al.} \cite{ladicky2014pulling} integrated depth estimation with semantic segmentation, and trained a classifier to solve these two problems.
The work in \cite{eigen2014depth} is among the pioneers to utilize CNNs to regress depth maps from a single view.
In their work, a global coarse-scale network is adopted to capture the overall scale, and then a local fine-scale network is adopted to refine the local details of the depth map.
The authors further trained their network on multiple tasks to demonstrate its generalization ability \cite{eigen2015predicting}.
By virtue of a pretrained ResNet50 and an efficient up-sampling scheme, Laina \textit{et al.} \cite{laina2016deeper} constructed a fully convolutional residual network, which decreases the evaluation errors by a large margin. Hu \textit{et al.} later extended \cite{laina2016deeper} in the feature extraction stage with squeeze-and-excitation blocks \cite{hu2018squeeze}. 
In line with the success of CNNs, Liu \textit{et al.} \cite{liu2014discrete} introduced a structural loss in the CNNs training, and Wang \textit{et al.} \cite{wang2015towards} further refined the estimated depth using a Hierarchical CRF.
More recently, Xu \textit{et al.} \cite{xu2017multi} proposed to integrate side outputs of the model by CRFs. Zhang \textit{et al.} \cite{zhang2018depth} trained the FCRN \cite{laina2016deeper} in a supervised manner under an adversarial learning framework. It is worth to note that, most monocular depth estimators either rely on large amounts of ground truth depth data or predict disparity as an intermediary step. To this end, Atapour-Abarghouei \textit{et al.} \cite{atapour2018real} trained a depth estimation model using pixel-perfect synthetic data, which overcomes the domain bias to resolve the above issues to a certain extent.

Since a high-quality depth map is not always easy to collect, unsupervised methods have also been introduced. Garg \textit{et al.} \cite{garg2016unsupervised} exploited the consistency between two registered views. They first warped the right image using the predicted inverse depth map, and then trained the estimation network to minimize the reconstruction error between the left image and the warped right image. Kumar \textit{et al.} \cite{cs2018monocular} followed the work in \cite{garg2016unsupervised} by using a discriminator network in an adversarial framework to distinguish the reconstructed image and the real image.
Xie \textit{et al.} \cite{xie2016deep3d} designed a similar pipeline and combined disparity maps of multiple levels to predict the right view, in which the objective function is the $L_1$ loss between the output right view and the ground truth right view.
Godard \textit{et al.} \cite{godard2017unsupervised} proposed a novel objective function that considers appearance matching loss, disparity smoothness loss and left-right disparity consistency loss, which achieves promising results in depth estimation. Requiring only a monocular video as input, Ranftl \textit{et al.} \cite{ranftl2016dense} reconstructed complex dynamic scene depth from temporal sequences by motion models.
Zhou \textit{et al.} \cite{zhou2017unsupervised} presented a supervised scheme for depth estimation using unlabeled video clips with view synthesis, which is implemented by a Depth CNN and a Pose CNN.

Besides the supervised and unsupervised solutions, semi-supervised depth estimation is also studied very recently.
Kuzniestsov \textit{et al.} \cite{kuznietsov2017semi} used a sparse ground truth depth map together with a registered stereo image pair to train a CNN, which has reached the state-of-the-art performance on the KITTI dataset \cite{Geiger2013IJRR}.
Note that our approach has a very different setting about the training data from \cite{kuznietsov2017semi}.
During training, we do not require all the image to have a registered depth (only a limit amount of image-depth pairs are needed), and we also need less ground truth depth maps to get comparable results, as quantitatively shown in Sec. 4.5. 

\section{The Proposed Approach}
\label{sec:method}


\subsection{Model Architecture}

The proposed generative adversarial learning framework consists of one generator network and two discriminator networks.
The generator network accepts an unannotated image as its input and predicts a corresponding depth map.
The predicted depth is then fed to: (1) a \emph{Pair Discriminator} (PD) network together with the RGB channels as its input;
(2) a \emph{Depth Discriminator} (DD) network together with a real depth map sampled from the labeled data, which gives feedback to the generator about whether the predicted depth comes from the same distribution of the real depth.
It is worth noting that, Nguyen \textit{et al.} \cite{nguyen2017dual} have used a similar architecture called D2GAN, but their generator is not conditioned and the two discriminators are designed to inspect the predicted data twice to avoid model collapse. D2GAN is unable to focus on the predicted depth map while guaranteeing its consistency with the corresponding RGB image, thus are not suitable for depth estimation task.

\begin{figure}
	\centering\includegraphics[width=1\linewidth]{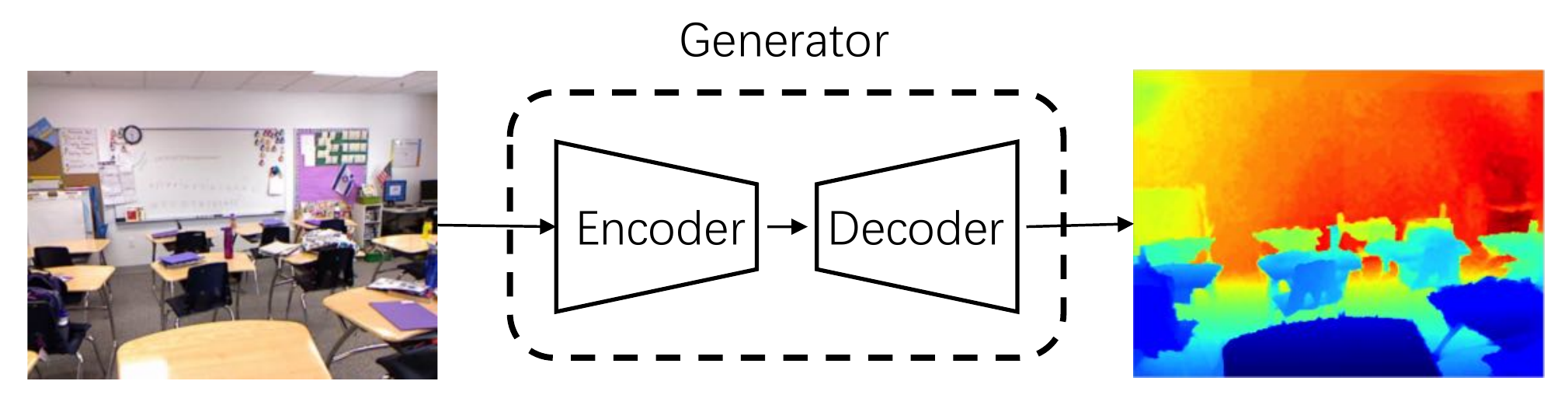}
	\caption{A basic encoder-decoder generator architecture.
    Encoder extracts features while reducing the spatial size, which is usually done by basic convolutions or convBlocks\cite{he2016deep,hu2018squeeze}. Decoder gradually upsamples the extracted features to a size similar with the input image, which is usually implemented by deconvolution or naive interpolation.}
	\label{fig:generator}
\end{figure}

\subsubsection{Generator Network}
\label{sec:generator}

Similar to semantic segmentation \cite{long2015fully}, edge detection \cite{xie2015holistically}, and other image translation tasks \cite{isola2017image}, depth estimation is typically conducted by using an encoder-decoder network, as illustrated in Fig. \ref{fig:generator}.
However, instead of classifying each pixel to a discrete label, the depth estimation network conducts a regression task that maps the pixel intensities to continuous depth values.
To accomplish such a dense prediction, most existing works \cite{eigen2014depth,eigen2015predicting,laina2016deeper,xu2017multi} firstly use a CNN to progressively extract feature maps from low level to high level.
After that, a decoder network is plugged in to regress the depth while restore the spatial resolution by upsampling the compact feature maps.  
To this end, Eigen \textit{et al.} \cite{eigen2014depth} proposed an architecture that contains two phases -- a coarse phase that produces low-resolution predictions using both convolution and fully-connected layers, and a fine phase that refines the first phase's output with more convolution layers.
Notably, the use of the fully-connected layer in the first phase enables the coarse network to have a full receptive field, which is essential to capture the global scale of the scene.
However, such a design is memory intensive given the large amount of parameters.
Laina \textit{et al.} \cite{laina2016deeper} took advantage of the ResNet architecture \cite{he2016deep} and designed a fast up-convolution module, which is similar to a ResNet block, but has a reverse data flow.
Hu \textit{et al.} \cite{hu2019revisiting} later inherited the decoding scheme, who adopts Squeeze-and-Excitation blocks \cite{hu2018squeeze} and intermediate feature fusion with an edge-aware loss function to train the model. 

In this work, the generator design can be of any cutting-edge depth estimation models (or more broadly image-to-image translation models). 
As quantitatively demonstrated later in our experiments, we have found that most of the modern architectures can be easily plugged into our semi-supervised framework as the generator to get a performance gain. However, unlike the previous works \cite{eigen2014depth,eigen2015predicting,laina2016deeper} that predict a smaller depth map of the corresponding RGB image, we predict the depth map with the same size as the input image.
And instead of reconstructing the original spatial resolution using naive methods such as bilinear interpolation, we want the network to learn the upsampling scheme by itself.
Throughout the paper, we use the network proposed in \cite{hu2019revisiting} as our generator backbone, and add an additional deconvolution layer at the end to upsample to the input resolution.
Furthermore, we also tested the models described in \cite{laina2016deeper,ronneberger2015u}, the corresponding comparison is shown later in Tab.~\ref{tab:model_improvement}.

\begin{figure}
	\centering\includegraphics[width=1\linewidth]{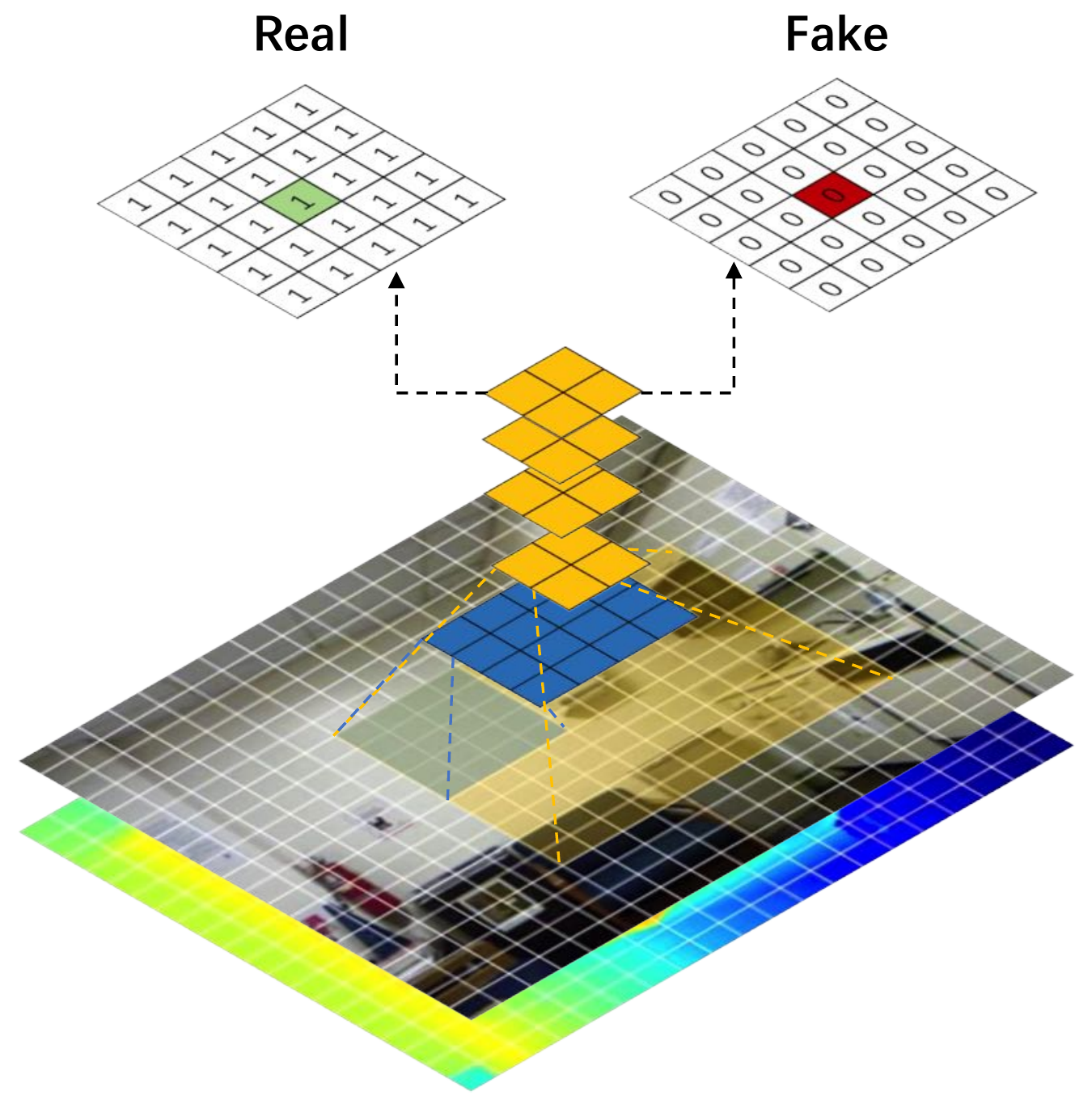}
	\caption{The architecture of the pair discriminator (PatchGAN\cite{isola2017image}), which consists of five layers of convolution with increasing sizes of receptive fields in each layer. The input image and the predicted depth map in the last layer are split to patches and each patch is inspected as real or fake by the discriminator.} 
	\label{fig:discriminator}
\end{figure}
\subsubsection{Discriminator Networks}

The two discriminator networks measure how far the distribution of the predicted depth map (and the image) is from the real one(s).
Naive losses such as $L_2$ norms penalize the generator by per-pixel discrepancy, which tends to lead overall blurry depth map either theoretically or empirically. The discriminator network's feedback (sometimes referred as GAN loss \cite{isola2017image}) as a contrast, distinguish such artifacts and penalize the generator to produce a depth map that accords with the natural depth value distribution, thus encourages more high frequencies (crispness/edge/detail) in the predicted depth map. The GAN loss performs similar effects to the higher-order potential of CRFs \cite{luc2016semantic}, since it can access the joint configuration of many depth patches. It helps to guide the generator to produce sharp depth maps and consider neighbor relations better than the pairwise potential.

For the pair discriminator, we use the PatchGAN \cite{isola2017image} to predict the \emph{real}/\emph{fake} labels at the scale of patches.
As shown in Fig. \ref{fig:discriminator}, the pair discriminator classifies each overlapped $70 \times 70$ patch in the input image-depth pairs to be \emph{real} or \emph{fake}, and computes a loss $D$ by averaging all responses.
Although the patch size $N$ can be of any size between $1$ and $\min(h_\text{input},w_\text{input})$, Isola \textit{et al.} \cite{isola2017image} have verified that a size around $70$ is best to encourage the overall sharpness while preventing artifacts at patch borders.
Isola \emph{et al.}~\cite{isola2017image} also showed that smaller or bigger patches do not improve performance on the task of semantic segmentation.
For our implementation, the patch sizes, \emph{a.k.a.} the receptive field sizes of each layer's neuron, are illustrated in Tab.~\ref{tab:receptivefield}.
The network has five convolution layers in total. 
The first four layers are followed by a batch normalization layer and a leaky ReLU layer to improve the stability.

For the depth discriminator, we use a network similar to the pair discriminator, with the difference that it only receives the predicted depth map and the real depth map as its inputs.
Other details of the model architecture are the same as the pair discriminator.

\begin{table}
    \caption{The receptive field size of neurons in each feature layer. A larger size takes more contextual information into account. }
	\label{tab:receptivefield}
	\centering
	\scriptsize
	\begin{tabular}{ c c c c c c}
		\hline
		Layer&1&2&3&4&5\\
		\hline
        Receptive field size & $4 \times 4$ & $10 \times 10$ & $22 \times 22$ & $46 \times 46$ & $70 \times 70$ \\
		\hline
	\end{tabular}
\end{table}

\subsection{Loss Function}
\label{subsec:lossfunc}

Assuming we have $n+m$ images ($n \ll m$) and $n$ depth maps of the first $n$ images, the training data we used can be expressed as:
\begin{displaymath}
\begin{aligned}
\mathbb{I} &= \{I_1,I_2,...,I_n,I_{n+1},...,I_{n+m}\}, \\ \mathbb{Y} &= \{Y_1,Y_2,...,Y_n\},
\end{aligned}
\end{displaymath}
where $I_i$ has a shape of $h \times w \times 3$ representing an RGB image, and $Y_i$ has a shape of $h \times w$ representing a depth map.
Our goal is to use all the $n+m$ images to learn a mapping $f$ from the image domain $I$ to the depth domain $Y$, by minimizing a loss function.
Different traditional GANs' loss, our loss function is tailored to the problem in two ways.

First the function is defined solely on the input image and output depth, without the noise component ${\mathbf z}$, which is different from the vanilla GAN \cite{goodfellow2014generative} and Conditional GANs \cite{mirza2014conditional}.
The vanilla GAN tries to map a random noise vector $\mathbf{z}$ to a desired image $\mathbf{y}$, which can be represented by $\mathbf{z} \stackrel{G}{\mapsto} \mathbf{y}$.
Conditional GANs \cite{mirza2014conditional} condition the mapping by adding image $\mathbf{i}$ to the generator, which can be viewed as $\{\mathbf{i},\mathbf{z}\}\stackrel{G}{\mapsto} \mathbf{y}$.
However, as observed in \cite{mathieu2015deep} and \cite{isola2017image}, the input noise vector $\mathbf{z}$ tends to be ignored by the generator and does not affect the output result $\mathbf{y}$ much.
Therefore, we discard the noise vector $\mathbf{z}$ and only use image $\mathbf{i}$ as the input to the generator, which is denoted by $\mathbf{i} \stackrel{G}{\mapsto} \mathbf{y}$ in depth estimation task($\mathbf{i}$ represents the image and $\mathbf{y}$ represents the predicted depth map). 

The second difference comes from the novel structure of the proposed GAN, having three components for the Generator, the Pairwise Discriminator, and the Depth Discriminator respectively.
The loss function for the Pair Discriminator is designed as:  
\begin{equation}
\label{eq:pdloss}
\begin{aligned}
\mathcal{L}_{PD}= & -\mathbb{E}_{i,y\thicksim p_\text{data}(i,y)} \Big[ \log PD(i,y) \Big] \\
&-\mathbb{E}_{i' \thicksim p_\text{data}(i')} \Big[\log\Big(1-PD\big(i',G(i')\big)\Big)\Big].
\end{aligned}
\end{equation}
Here $PD(x,y)$ represents the probability that $(x,y)$ is a real pair from the training set.
$i$ is sampled from $\{I_1,I_2,...,I_n\}$ and $i'$ is sampled from $\{I_{n+1},...,I_{n+m}\}$.
The depth map is stacked with the corresponding image to allow the discriminator to penalize the mismatch between the joint distribution of $(i,d)$ and $(i',G(i'))$.

In order to predict high-quality depth maps, the generator should fool the pair discriminator by making $PD(i',G(i'))$ as close to $1$ as possible.
The corresponding objective for G is then:
\begin{equation}
\label{eq:gloss}
\mathcal{L}_{G}^{PD}=\mathbb{E}_{i' \thicksim p_\text{data}(i')} \Big[ \log \Big( 1-PD \big(i',G(i') \big) \Big) \Big].
\end{equation}
Note that maximizing $-\mathbb{E}_{i' \thicksim p_\text{data}(i')}\Big[\log \Big(1-PD\big(i',G(i') \big)\Big)\Big]$ means to maximizing the probability that $D$ makes a mistake, which is equalized to minimizing $\mathbb{E}_{i' \thicksim p_\text{data}(i')} \Big[ \log \Big(1-PD\big(i',G(i') \big) \Big) \Big]$.

The depth discriminator (DD), however, only looks at the predicted map to infer whether the generated map is drawn from the same distribution with a limited amount of ground truth depth maps.
Its objective function is therefore written as:
\begin{equation}
\label{eq:ddloss}
\begin{aligned}
\mathcal{L}_{DD}=&-\mathbb{E}_{y\thicksim p_\text{data}(y)}\Big[\log DD(y) \Big] \\ 
&-\mathbb{E}_{i' \thicksim p_\text{data}(i')} \Big[ \log \Big(1-DD \big(G(i') \big) \Big) \Big].
\end{aligned}
\end{equation}
Correspondingly, the generator computes the loss by using the following equation:
\begin{equation}
\mathcal{L}_{G}^{DD}=\mathbb{E}_{i' \thicksim p_\text{data}(i')}\Big[\log\Big(1-DD\big(G(i')\big)\Big)\Big].
\end{equation}
Note that the depth discriminator does not care if the generated depth is correspondent to the image.
It only aims to ensure that the predicted map looks ``natural'', \emph{i.e.} the map is not distinguishable from the sample drawn from $p_\text{data}(d)$.

Finally, we combine the two discriminator networks with the generator network.
More formally, PD, DD and G play a three-player minimax optimization game as below:
\begin{equation}
\begin{aligned}
\min_G\max_{PD,DD}\mathcal{V}(G,PD, &DD)=\mathbb{E}_{i,y\thicksim p_\text{data}(i,y)} \Big[ \log PD(i,y) \Big]\\ +
& \mathbb{E}_{i' \thicksim p_\text{data}(i')} \Big[ \log \Big( 1-PD\big(i',G(i') \big) \Big) \Big] \\ +
& \mathbb{E}_{y\thicksim p_\text{data}(y)}\Big[\log PD(d) \Big] \\ +
& \mathbb{E}_{i' \thicksim p_\text{data}(i')} \Big[ \log \Big( 1-PD \big( G(i') \big) \Big) \Big].
\end{aligned}
\end{equation}
The pair discriminator and depth discriminator is optimized using Eqs. (\ref{eq:pdloss}) and (\ref{eq:ddloss}) respectively. 
And the generator now considers feedbacks from these two discriminators:
\begin{equation}
\begin{aligned}
\mathcal{L}_{G}=&\lambda\mathbb{E}_{i' \thicksim p_\text{data}(i')} \Big[ \log \Big(1-PD\big(i',G(i') \big) \Big) \Big] \\+ 
&(1-\lambda)\mathbb{E}_{i' \thicksim p_\text{data}(i')} \Big[ \log \Big(1-DD \big(G(i') \big) \Big) \Big],
\end{aligned}
\end{equation}
where $0<\lambda\leq1$ is a hyperparameter to control the trade-off between the pair discriminator and the depth discriminator. 
In the experiment, we have found that $\lambda=0.7$ gives a reasonable result both qualitatively and quantitatively, which means the feedback from the pair discriminator should be given more attention. 
Following the work in \cite{isola2017image}, we alternatively optimize PD, DD and G one step at a time.

\begin{figure*}
	\centering\includegraphics[height=2.7in]{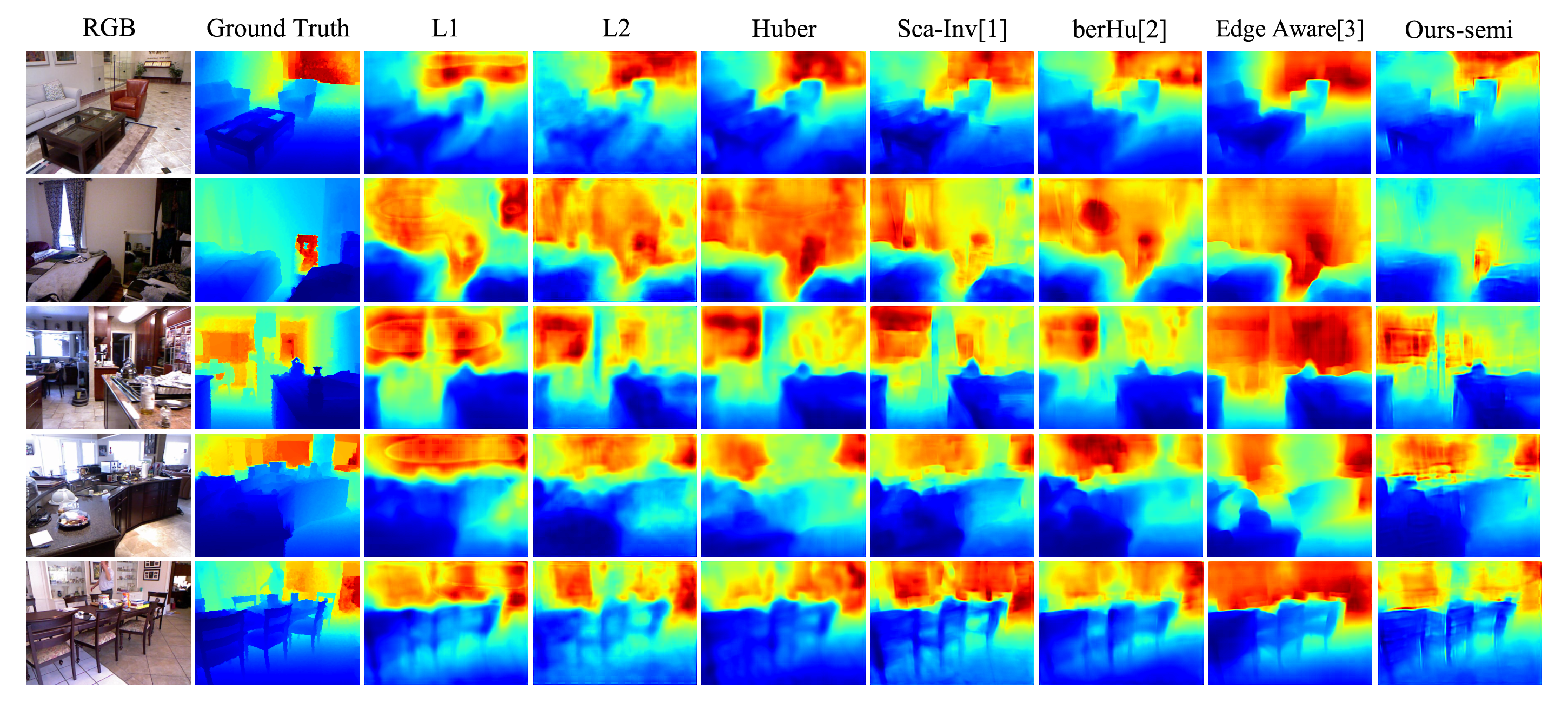}
	\caption{Qualitative results on NYU Depth test set. All losses are applied to the same model architecture with the same learning strategy using 500 image-depth pairs.
    Our semi-supervised models are able to produce finer  depth maps than the compared approaches.}
	\label{fig:nyud}
\end{figure*}

\subsection{Training Details}
\label{subsec:training_details}
We embed the semi-supervised training phase into the supervised training phase. During training, supervised training and semi-supervised training switches between iterations. In the supervised iteration, we make full use of the ground truth depth map by computing both regression loss $L_1$ and GAN loss as follows:
\begin{equation}
\begin{aligned}
\mathcal{L}_{G}=&\lambda\mathbb{E}_{i \thicksim p_\text{data}(i)} \Big[ \log \Big(1-PD\big(i,G(i) \big) \Big) \Big] \\+ 
&(1-\lambda)\mathbb{E}_{i \thicksim p_\text{data}(i)} \Big[ \log \Big(1-DD \big(G(i) \big) \Big) \Big]\\+
&\beta \mathbb{E}_{i,y \thicksim p_\text{data}(i,y)} \Big[ \|y-G(i) \|_1\Big],
\end{aligned}
\end{equation}
where parameters $\lambda$ and $\beta$ control the weights between the regression loss and the GAN loss. In our experiment, we have found that a larger $\beta$ at the initial training gives better results. The parameters of pair discriminator and depth discriminator are updated by computing the loss as below:
\begin{equation}
\begin{aligned}
\mathcal{L}_{PD}= & -\mathbb{E}_{i,y\thicksim p_\text{data}(i,y)} \Big[ \log PD(i,y) \Big] \\
&-\mathbb{E}_{i \thicksim p_\text{data}(i')} \Big[\log\Big(1-PD\big(i,G(i)\big)\Big)\Big].
\end{aligned}
\end{equation}
\begin{equation}
\begin{aligned}
\mathcal{L}_{DD}= & -\mathbb{E}_{i,y\thicksim p_\text{data}(i,y)} \Big[ \log PD(i,y) \Big] \\
&-\mathbb{E}_{i \thicksim p_\text{data}(i)} \Big[\log\Big(1-PD\big(i',G(i')\big)\Big)\Big].
\end{aligned}
\end{equation}
In the semi-supervised phase, we optimize the generator and discriminator networks by the loss functions defined in Sec. \ref{subsec:lossfunc}.
Note that the key difference between the supervised training phase and the semi-supervised training phase is that all the losses are computed by labeled image-depth pairs without using the unlabeled images in the supervised training phase.
We have found that it can stabilize the training process to get the best performance.

\section{Experiments}
\label{sec:exp}
We conduct our experiments in three aspects:
\begin{itemize}
\item{In Sec. ~\ref{subsec:NyuExp}, we first validate that the proposed GAN loss is superior in the depth regression problem over other losses using a limited amount of ground truth data. 
Then we show that most of the cutting-edge models can get a performance gain under our semi-supervised settings.
Last but not least, we conduct extensive experiments to explore when the semi-supervised framework can potentially improve the performance.}
\item{In Sec. ~\ref{subsec:make3d}, we show that our GAN loss can reach state-of-the-art performance on the Maked3D dataset, which verifies the point that our method has an advantage when the data amount is practically small.}
\item{In Sec. ~\ref{subsec:da}, we further testify the proposed method for the task of the domain adaptation, which has shown that the proposed semi-supervised scheme performs better than the traditional supervised schemes.}
\end{itemize}
We implement our model using PyTorch on NVIDIA GeForce GTX TITAN X GPU with 12GB GPU memory.
For the pair discriminator and the depth discriminator, we initialize parameters of all layers with values drawn from a normal distribution ($\mu=0$, $\sigma=0.02$). As for the generator, we warm it up by training it with a simple $L_1$ loss for initialization. The overall training pipeline is introduced in Alg. 1. And we have quantitatively observed that the pretrained initialization performs better than the normal initialization. We stop training when the loss of the generator has converged.

\subsection{Evaluation Protocols}
\label{sec:eva_protocols}
In alignment with \cite{eigen2014depth}, we evaluate our model with the following error metrics:
\begin{tabular}{l} 
			$ARD = \frac { 1} { | T | } \sum _ { y \in T } | y - y ^ { * } | / y ^ { * } $, 
			\cr$RMSE = \sqrt { \frac { 1} { | T | } \sum _ { y \in T } (y _ { i } - y _ { i } ^ { * })  ^ { 2} }$, 
			\cr$RMSE(log)= \sqrt { \frac { 1} { | T | } \sum_{y \in T} ( \log y _ { i } - \log y _ { i } ^ { * } ) ^ { 2} }$,  
			\cr$\log_{10} =  \frac { 1} { | T | } \sum _ { y \in T } |\log_{10} y_{i} - \log_{10} y_{i}^{*}|$, 
			\cr$\delta = \%\, of\, y_i\, s.t. \max \left( \frac { y _ { i } } { y _ { i } ^ { * } } ,\frac { y _ { i } ^ { * } } { y _ { i } } \right) < \text{ thr }$. 
\end{tabular}

In the equations above, ARD and RMSE are abbreviations for Absolute Relative Difference (rel) and Root Mean Squared Error, respectively. 
$y_i$ is the predicted pixel depth value and $y^*$ is ground truth depth value.
$T$ represents the pixel number in the test set.
We use $\delta_1$, $\delta_2$, and $\delta_3$ to denote the situations where $\delta<1.25$, $\delta<1.25^2$, and $\delta<1.25^3$. Note that for rel, RMSE, RMSE ($\log$) and $\log_{10}$ metrics, lower is better.
And, higher performance can be expected by setting a higher $\delta_i$ value.

\subsection{Baselines}
As mentioned in the previous sections, our framework can benefit most of the cutting-edge models.
However, model architecture itself sometimes greatly influences the estimation accuracy.
To show that our performance gain is general enough to benefit a wide range of backbone networks, we use off-the-shelf models \cite{ronneberger2015u,laina2016deeper,hu2019revisiting} as our baselines, which meet our assumptions for the generator network as  described in Sec. ~\ref{sec:generator}.

\begin{table*}
	\caption{Evaluation of different loss functions on NYU Depth test set.
    All the models are trained using the same generator architecture.
    Our GAN loss gives the best quantitative results when the number of training data is limited.}
	\label{tab:loss}
	\centering
	\begin{tabular}{ c c c c c c c c c c}
		\hline
		loss&type&\#gt used&rel&rms&$log_{10}$&$\delta_1$&$\delta_2$&$\delta_3$\\
		\hline
		$L_1$&supervised&500&0.201&0.750&0.083&0.680&0.917&0.978 \\
		$L_2$&supervised&500&0.195&0.706&0.080&0.695&0.924&0.981\\
		Huber&supervised&500&0.204&0.698&0.080&0.696&0.920&0.977\\
		Scale-invariant \cite{eigen2014depth}&supervised&500&0.192&\textbf{0.685}&0.078&0.712&0.929&0.983\\
		berHu \cite{laina2016deeper}&supervised&500&0.199&0.705&0.079&0.708&0.919&0.978\\
		Edge Aware \cite{hu2019revisiting}&supervised&500&0.201&0.750&0.083&0.681&0.917&0.978\\
		Ours (PD only)&semi-supervised&500&0.198&0.721&0.081&0.700&0.918&0.978\\
		Ours (DD only)&semi-supervised&500&0.191&0.708&0.078&0.709&0.925&0.980\\
		Ours&semi-supervised&500&\textbf{0.183}&0.704&\textbf{0.077}&\textbf{0.713}&\textbf{0.931}&\textbf{0.984}\\
		\hline
	\end{tabular}
\end{table*}

\subsection{NYU Depth Dataset}
\label{subsec:NyuExp}

NYU Depth v2 dataset \cite{silberman2012indoor} contains RGB images and the corresponding depth maps of various indoor scenes, which are captured by Microsoft Kinect v1.
The full training set contains $464$ scenes, which are officially split to $249$ scenes for training and $215$ scenes for testing.
Together with the full set, a sufficiently labeled subset is also offered.
This subset has $795$ training images and $654$ testing images.
Following the previous works in \cite{eigen2014depth,laina2016deeper}, we test on the $654$ images in the following experiments.
The raw training depth maps given by the dataset are in a range of $(0m,10m)$ and are stored as uint8 PNG images.
Before feeding the data to the generator network, we first downsample both the image and the depth map to a size of $320\times 240$ (half of the full resolution) in order to accelerate the training process, each of which is then center-cropped to a size of $304 \times 228$ to exclude the blank frame border.
The RGB images are normalized using the mean and standard deviation values computed on ImageNet \cite{russakovsky2015imagenet}.
Finally, we train our model for $20$ epochs with a batch size of $8$.
Parameters of the generator and the discriminator are all optimized by Adam \cite{kingma2014adam}, with an initial learning rate of $0.0002$ that is multiplied by 0.1 every $100$ iterations.
The coefficients $\beta_1$ and $\beta_2$ used for running the averages of gradient and its square are set to $0.9$ and $0.999$, respectively.

\begin{figure}
	\centering\includegraphics[width=3.2in]{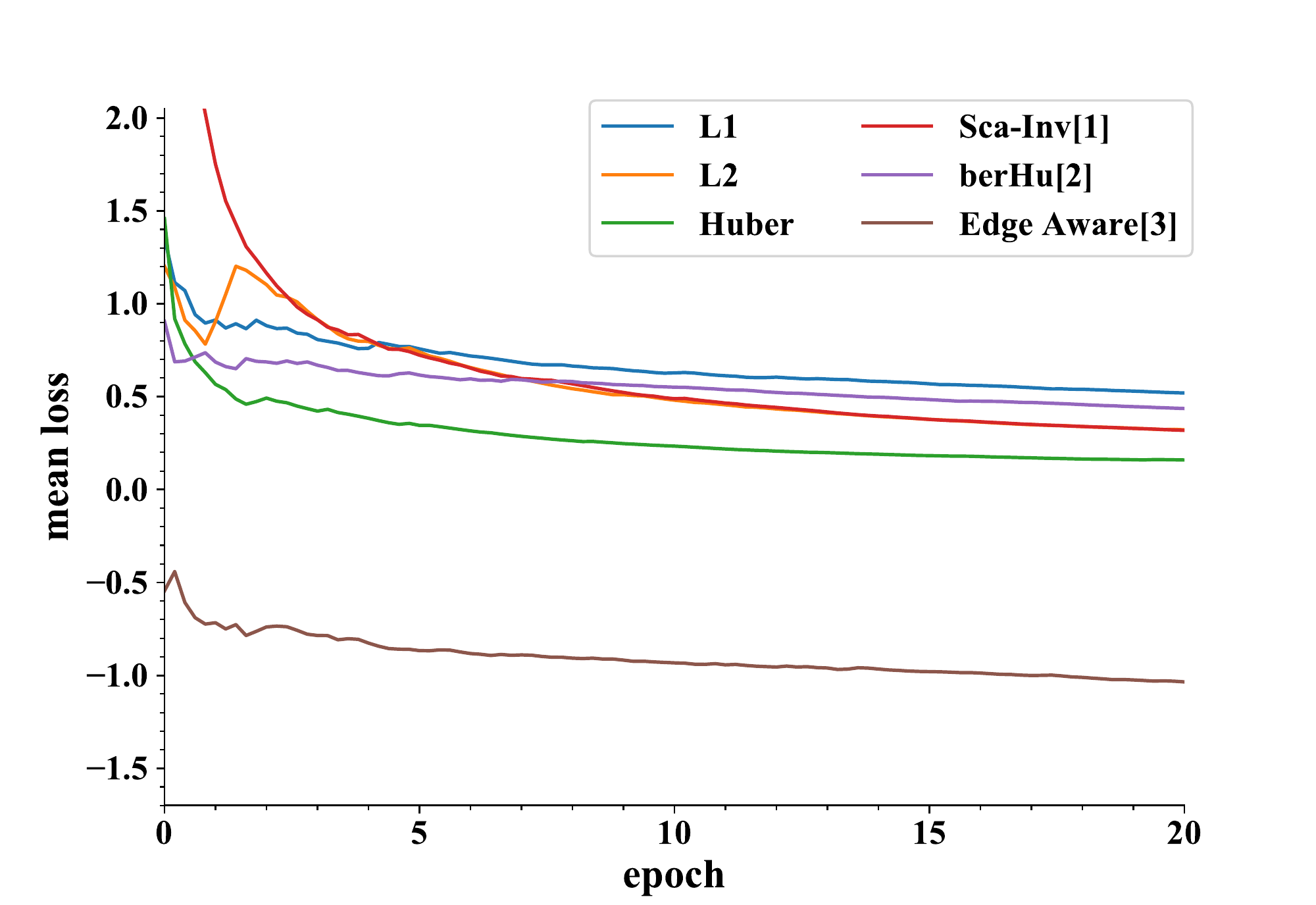}
	\caption{The convergence curves of different loss functions during model training. After about 15 epochs, all losses tend to stop decreasing and we stop training when convergence is observed.}
	\label{fig:loss_convergence}
\end{figure}

\textbf{Evaluating the loss functions.}
We first evaluate whether the proposed GAN loss can improve the overall performance.
We train our baseline generator with $L_1$, $L_2$, Huber, Sca-Inv\cite{eigen2014depth}, berHu\cite{laina2016deeper} and edge-aware \cite{hu2019revisiting} losses in a supervised manner, and with the loss proposed in Sec. ~\ref{subsec:lossfunc} in a semi-supervised manner. To study the effect of the discriminator structure, we experiment three types of the discriminator - pair discriminator only, depth discriminator only and a combination of the both.
We use $500$ labeled images to evaluate the predicted maps with measurements described in Sec. \ref{sec:eva_protocols}.
To make sure the losses are fairly treated during training, we plot the convergence curve in Fig. ~\ref{fig:loss_convergence} and only stop training when a loss convergence is observed.
Note that the loss of berHu is scaled by 0.1 for better visualization.
Quantitative results are reported in Tab.~\ref{tab:loss}. 
Among the nine losses, our semi-supervised loss outperforms all the other losses that are popularly used in the depth estimation task and is better than using any of the discriminators alone. We also randomly visualize some results to get a qualitative  sense of the predicted depth map. The visualizations are shown in Fig. \ref{fig:nyud}. It can be seen that $L_1$, $L_2$ and Huber losses are good at capturing the overall mean depth scale, but are prone to output blurry depth maps.
Scale-invariant error, instead, measures the scene point relationships that are irrespective to the absolute global scale, which can better preserve the relative depth.
Besides, berHu \cite{laina2016deeper} shows a good quantitative result considering that the depth distribution in NYU Depth dataset is heavily-tailed \cite{roy2016monocular}.
Our semi-supervised framework produces the best result, which preserves more sharpness at object borders, and are more natural than other losses. We also compare our method with \cite{zhang2018depth}, in which a GAN variant loss was proposed to train a depth estimator. Results are given in Tab. \ref{tab:compare_with_zhang}. We can see that though our semi-supervised framework is designed for situations when training data is limited, we could still outperform \cite{zhang2018depth} when training data is sufficient (12K in this case). More qualitative results are shown in Fig. \ref{fig:more_results} (a). 
We can see that the predicted maps give better results for the near objects, but sometimes tend to confuse the middle distance and far distance. 
This is actually also the case for human beings.
For near objects, we can even estimate the distance by a precision of centimeter.
But for far objects, it's very hard to get meter-level precision.
\begin{table}
	\caption{Comparisons with the GAN loss variant. Our method, though not designed for depth estimation when training data is sufficient (12K in this case), still performs better w.r.t most of the evaluation metrics.}
	\label{tab:compare_with_zhang}
	\begin{tabular}{p{2.0cm} p{1.3cm} p{0.7cm} p{0.7cm} p{1.0cm} p{0.7cm}}
			\hline
			algorithm&\#gt used&rel&rms&rms $log$&$\delta_1$\\
			\hline
			Zhang \textit{et al.}\cite{zhang2018depth}&12K&0.128&0.551&\textbf{0.170}&0.824\\
			Ours&12K&\textbf{0.124}&\textbf{0.549}&0.178&\textbf{0.851}\\
			\hline
	\end{tabular}
\end{table}

\begin{center}
	\begin{figure*}
	\centering\includegraphics[width=\linewidth]{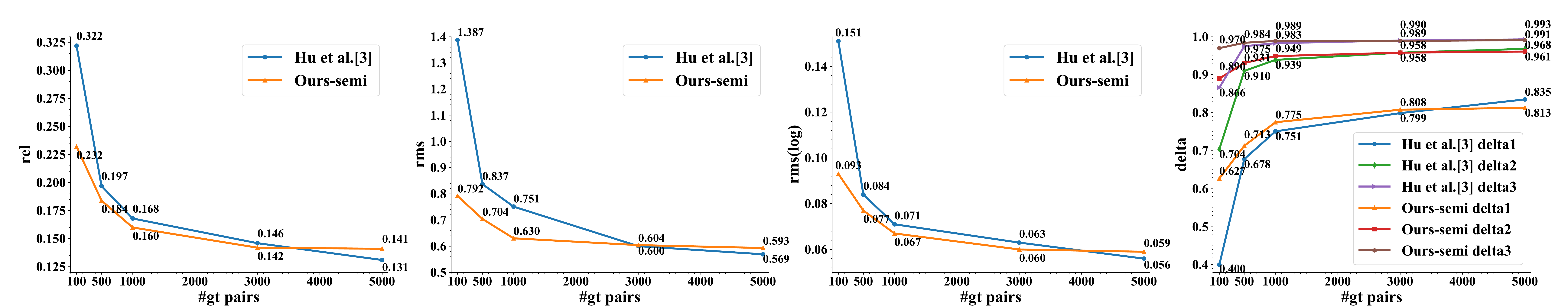}
    \caption{Performance curves with respect to different numbers of training image-depth pairs. We can see that our semi-supervised framework can boost the performance when the number of labeled training samples is less than 1K.}
	\label{fig:nyud_comp}
\end{figure*}
\end{center}

\textbf{Improvements over other methods.} Besides our baseline model of \cite{hu2019revisiting}, we also tried two different models \cite{ronneberger2015u,laina2016deeper} under our semi-supervised learning framework and compare the results with the ones that are trained by using the regression loss $L_1$. Comparison results are reported in Tab.~\ref{tab:model_improvement}. All three model architectures receive performance gains at different levels of data numbers. Further, to suggest a data range about when the proposed semi-supervised framework can bring performance gain, we train our baseline model with the loss that is also proposed in \cite{hu2019revisiting} in a supervised manner, and compare the results with ours in Fig. \ref{fig:nyud_comp}. It can be seen that when the training data is practically small (for example, around $100$ pairs), our methods can vastly boost the prediction accuracy. 
The margin gradually decreases when the training data becomes sufficient.
When there are more than $3.5$K training image-depth pairs, fully-supervised approaches are more suitable for the problem.

\begin{table}
	\caption{Quantitative comparisons of models trained by supervised learning and our semi-supervised framework.}
	\label{tab:model_improvement}
	\begin{tabular}{p{3.2cm} p{1.3cm} p{0.7cm} p{0.7cm} p{0.7cm}}
			\hline
			generator&\#gt used&rel&rms&$log_{10}$\\
			\hline
			Unet\cite{ronneberger2015u}&100&0.374&1.641&0.209\\
			Unet\cite{ronneberger2015u} semi&100&\textbf{0.367}&\textbf{1.264}&\textbf{0.146}\\
			Unet\cite{ronneberger2015u}&500&0.372&1.179&0.141\\
			Unet\cite{ronneberger2015u} semi&500&\textbf{0.369}&\textbf{1.154}&\textbf{0.137}\\
			Unet\cite{ronneberger2015u}&1000&0.365&1.180&0.137\\
			Unet\cite{ronneberger2015u} semi&1000&\textbf{0.369}&\textbf{1.141}&\textbf{0.135}\\
			\hline
			FCRN \cite{laina2016deeper}&100&0.286&0.964&0.114\\
			FCRN \cite{laina2016deeper} semi&100&\textbf{0.279}&\textbf{0.925}&\textbf{0.112}\\
			FCRN \cite{laina2016deeper}&500&0.272&0.906&0.111\\
			FCRN \cite{laina2016deeper} semi&500&\textbf{0.254}&\textbf{0.849}&\textbf{0.102}\\
			FCRN \cite{laina2016deeper}&1000&0.220&0.760&0.086\\
			FCRN \cite{laina2016deeper} semi&1000&\textbf{0.210}&\textbf{0.736}&\textbf{0.083}\\
			\hline
			Hu et. al. \cite{hu2019revisiting}&100&0.322&1.387&0.151\\
			Hu et. al. \cite{hu2019revisiting} semi&100&\textbf{0.232}&\textbf{0.792}&\textbf{0.093}\\
			Hu et. al. \cite{hu2019revisiting}&500&0.197&0.837&0.084\\
			Hu et. al. \cite{hu2019revisiting} semi&500&\textbf{0.184}&\textbf{0.704}&\textbf{0.077}\\
			Hu et. al. \cite{hu2019revisiting}&1000&0.168&0.751&0.071\\
			Hu et. al. \cite{hu2019revisiting} semi&1000&\textbf{0.160}&\textbf{0.630}&\textbf{0.067}\\
			\hline
	\end{tabular}
\end{table}
\begin{table}
	\caption{Error analysis against different semantic areas.}
	\label{tab:semantic}
	\begin{tabular}{p{1.1cm} p{0.8cm} p{0.8cm} p{0.8cm} p{0.8cm} p{0.8cm} p{0.8cm}}
		\hline
		\hfil area&\hfil rel&\hfil rms&$\hfil log_{10}$&\hfil $\delta_1$&\hfil $\delta_2$&\hfil $\delta_3$\\
		\hline
		floor&\textbf{0.172}&\textbf{0.431}&\textbf{0.072}&\textbf{0.745}&\textbf{0.938}&\textbf{0.986}\\
		structure&0.192&0.831&0.081&0.690&0.919&0.981\\
		props&0.187&0.694&0.078&0.711&0.931&0.984\\
		furniture&0.176&0.612&0.074&0.728&0.938&0.987\\
		missing&0.186&0.721&0.078&0.708&0.930&0.983\\
		\hline
		overall&0.183&0.704&0.077&0.713&0.931&0.984\\
		\hline
	\end{tabular}
\end{table}

\textbf{Error analysis.}
We further analyze the prediction errors with respect to different semantic regions.
Besides images and depth maps, the NYU Depth v2 dataset also provides semantic labels for each image.
Silberman \textit{et al.} \cite{silberman2012indoor} defined a general 4-class labels (\emph{floor}, \emph{structure}, \emph{props} and \emph{furniture}) plus one missing area out of the original 894 class labels.
We follow this class combination strategy and compute the error in these five areas.
Results are reported in Tab.~\ref{tab:semantic}. 
We can see that our model performs best in the \emph{floor} area and performs worst in the \emph{structure} area (ignoring the label missing area). 
It may be due to the fact that floors are smoother with less depth hop, while the depth of structure is more protean and often shows in far distance. 
\begin{figure}
	\centering\includegraphics[width=\linewidth]{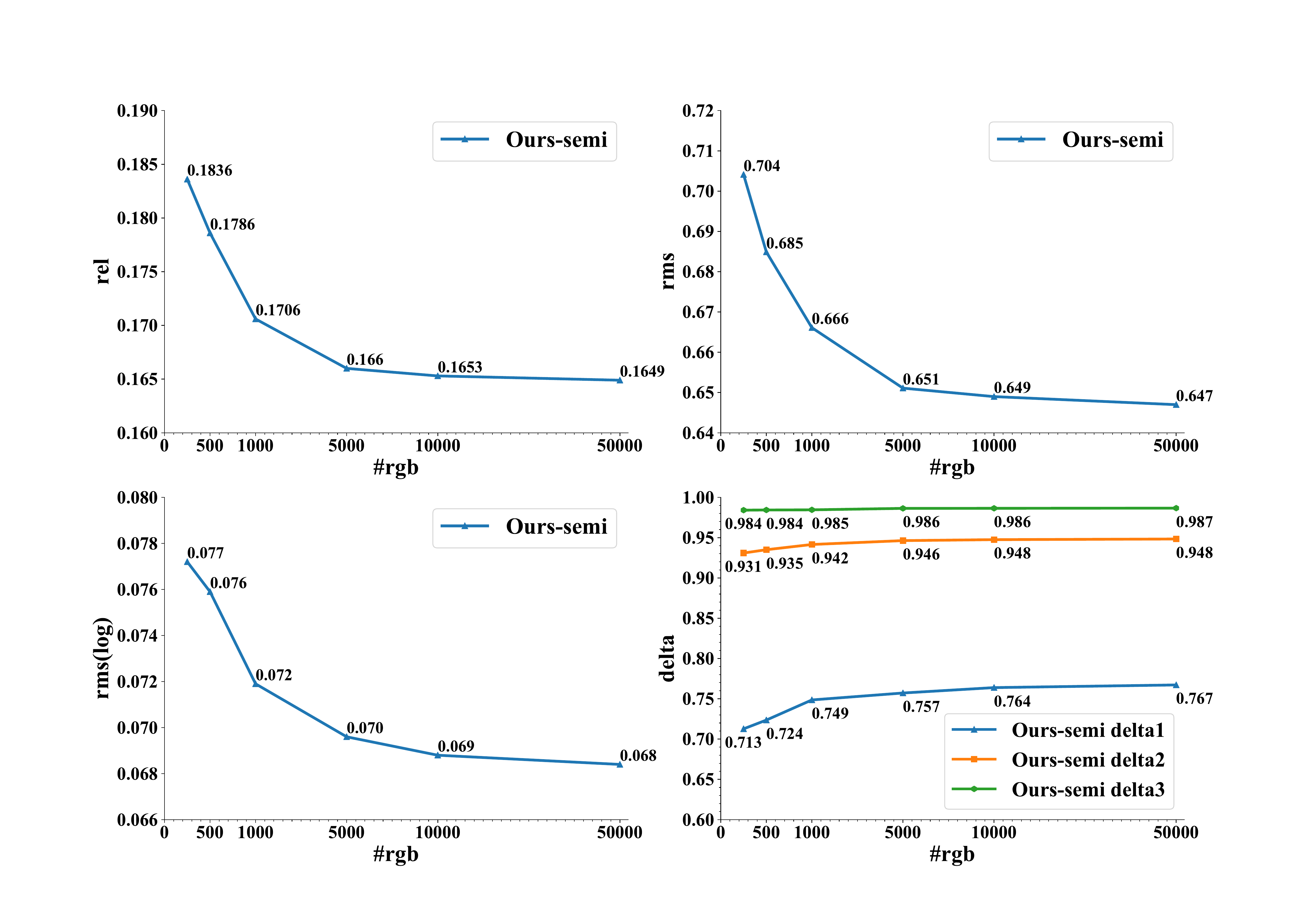}
	\caption{Performance curves with respect to different numbers of additional RGB images used on NYU Depth dataset. In all experiment, 500 image-depth pairs are used. Our semi-supervised framework can effectively boost the performance by leveraging extra unlabeled RGB images.}
	\label{fig:additional_RGB}
\end{figure}
\begin{figure}
	\centering\includegraphics[width=\linewidth]{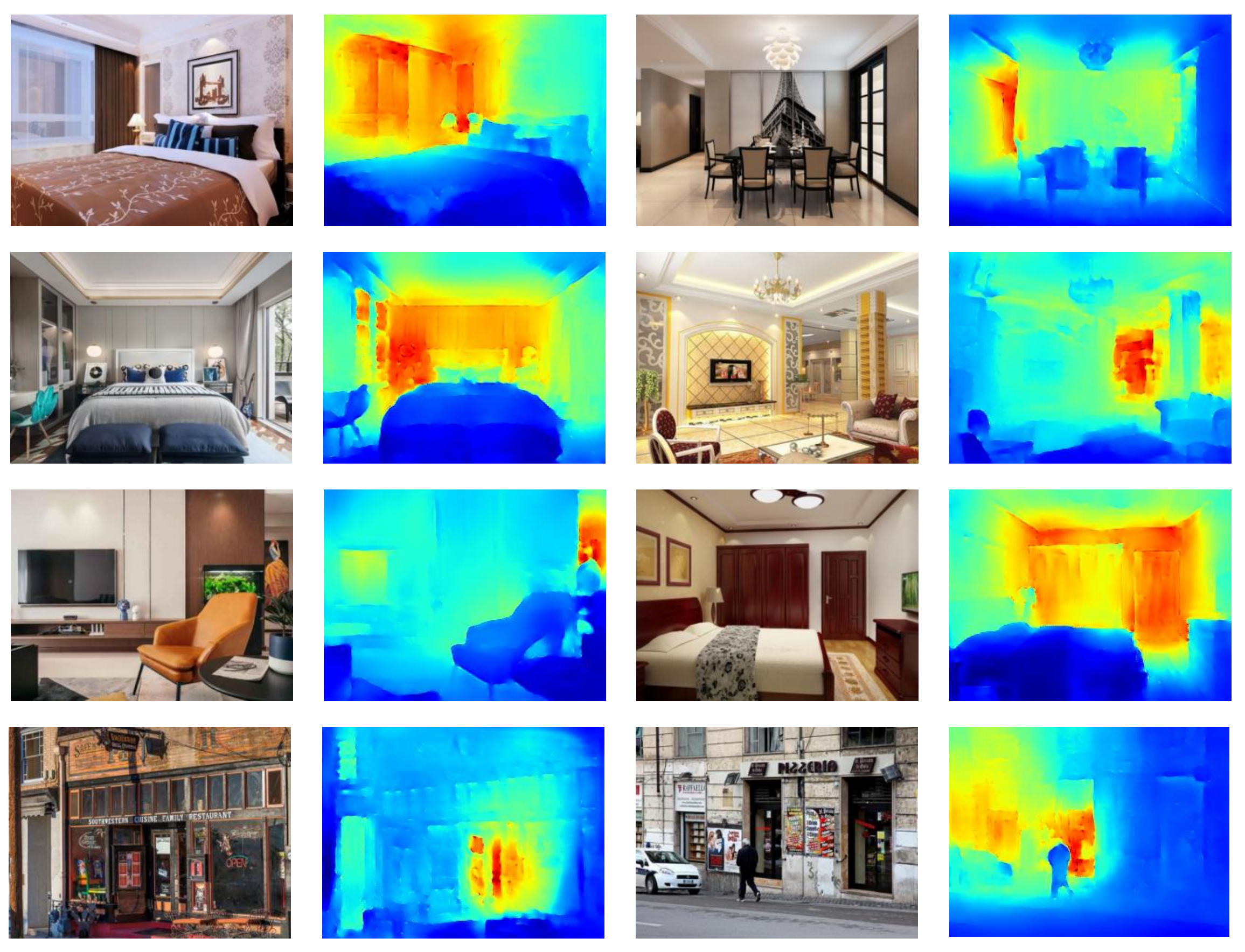}
	\caption{Some qualitative results tested on wild images using the model trained on NYUDv2 with 1k ground truth image pairs.}
	\label{fig:res_in_wild}
\end{figure}

\textbf{Ablation study towards the additional RGB image number.} To further evaluate the effect of the additional RGB number on our semi-supervised NYU Depth model performance, we experiment $500$, $1k$, $5k$, $10k$ and $50k$ RGB images respectively together with $500$ image-depth pairs in our semi-supervised setting. Results are given in Fig. \ref{fig:additional_RGB} , from which we can see that 
given $k$ training image-depth pairs, adding more additional RGB images (within a range of $[1/2*k,10*k]$) is beneficial to the model performance. The performance gain gradually saturates after 10 times more RGB images are used.

\textbf{Testing on images in the wild.}
We randomly pick some Internet images covering both indoor and outdoor scenes, and then test our model on them to see whether the model generalizes well to unseen scenes drawn from an unfamiliar data distribution. Results are shown in Fig. \ref{fig:res_in_wild}. We compare the predicted depth with our visual estimation and found that the result is decent.
For the indoor scenes, our model often predicts depth with a high accuracy.
For the outdoor scenes,  our model predicts relative depth decently but fails to capture the absolute distance scale.
It can be explained by domain shift, since the absolute distance of outdoor scenes is significantly different from the NYU Depth indoor dataset.
In most cases, the model predicts reliable and convincing results, which is useful for various applications, such as cleaning robots, UAVs, and computational photography.

\subsection{Make3D Dataset}
\label{subsec:make3d}
Make3D \cite{saxena2009make3d} is an outdoor dataset captured by an RGB camera and a laser scanner.
It contains $534$ aligned image and range data, which are officially divided into $400$ training pairs and $134$ testing pairs.
The RGB images are stored with a $2,272 \times 1,704$ resolution, while their depth maps are stored with a $55 \times 305$ resolution, which is an order of magnitude smaller in the spatial size. Thus, we first resize all images and the corresponding depth maps to a uniform size of $320 \times 240$ using bilinear interpolation.
Besides, due to the inaccurate long-range distance and glasses in data acquisition, we mask out all invalid depth pixels and pixels with distance $>$ 70 meters following \cite{laina2016deeper}.
We first train our model with all training image-depth pairs in a supervised manner as described in Sec. \ref{subsec:training_details} to get a fair comparison with other supervised methods. Then, we take another 425 RGB images from a similar dataset, namely the \textit{dataset-2} of the Make3D dataset, to further evaluate the effectiveness of the proposed semi-supervised learning.

\begin{figure}
	\centering\includegraphics[width=3.5in]{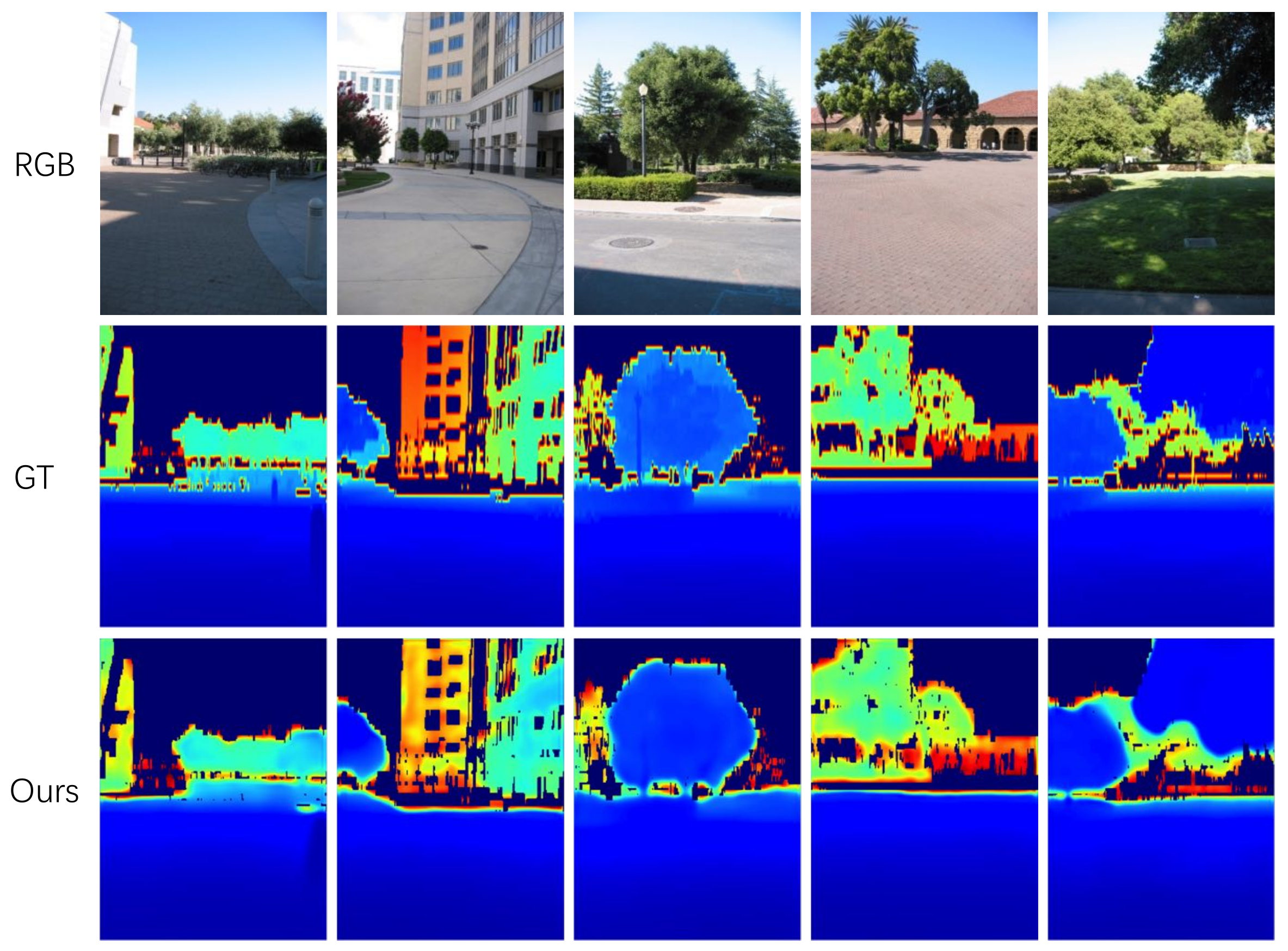}
	\caption{Qualitative results on Make3D dataset.
    The rows (from up to bottom) are RGB images, ground truth depth maps, and results by our supervised model, respectively. Pixels with distance larger than $70m$ are masked out.}
	\label{fig:qmake3d}
\end{figure}

\begin{figure*}
	\centering\includegraphics[width=5.7in]{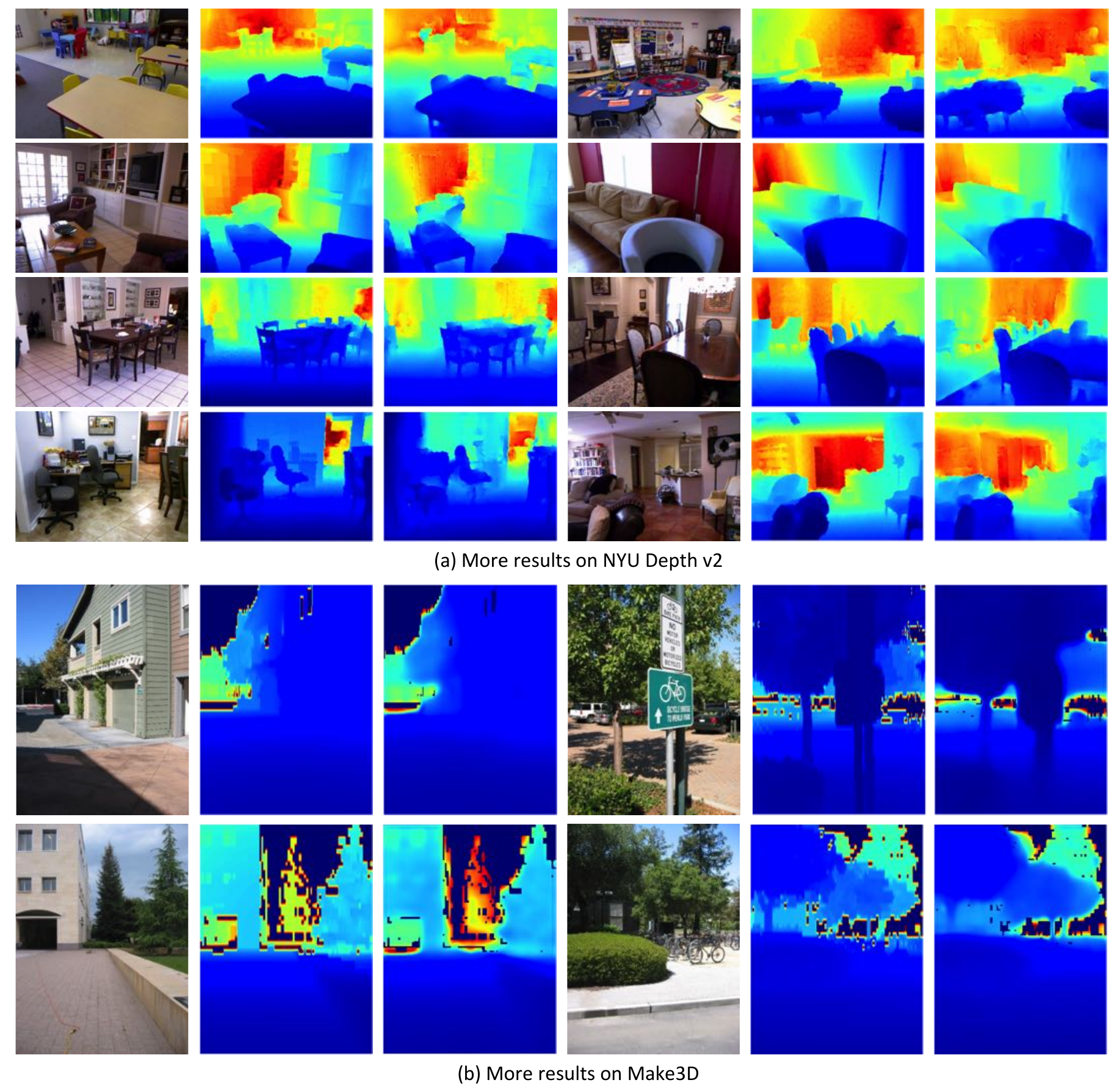}
	\caption{More qualitative results on NYUD dataset and Make3D dataset.
        The columns from left to right are RGB images, ground truth depth map and predicted depth by our model.
    Depth values are normalized for visualization.}
	\label{fig:more_results}
\end{figure*}

We compare our methods with \cite{karsch2012depth,liu2014discrete,laina2016deeper,kuznietsov2017semi} and report the results in Tab.~\ref{tab:make3d}. Our models push the evaluation errors down to a new level and the semi-supervised scheme can reach the state-of-the-art performance on all metrics. We can thus conclude that the semi-supervised learning effectively utilizes the extra RGB images and can generalize well to outdoor scenes. 
Sample qualitative results are shown in Fig.~\ref{fig:qmake3d} and more qualitative results are given in Fig. \ref{fig:more_results} (b). It can be observed that our model overcomes the distraction from the shadow area and can capture the underlying structure behind the scene appearance image well.
\begin{table}
	\caption{Comparison on the Make3D dataset. Our methods generalize well to outdoor scenes and can reach state-of-the-art performance. Pixels with distance larger than $70m$ are masked out.}
	\label{tab:make3d}
	\scalebox{0.8}[0.8]{
		\begin{tabular}{p{3.2cm} p{2.3cm} p{0.7cm} p{0.7cm} p{0.7cm}}
			\hline
			algorithm&type&rel&rms&$log_{10}$\\
			\hline
            Make3D \cite{saxena2009make3d}&supervised&0.370&-&0.187\\
            Liu \textit{et al.} \cite{liu2010single}&supervised&0.379&-&0.148 \\
            DepthTransfer \cite{karsch2014depth}&supervised&0.361&15.10&0.148\\
            Liu \textit{et al.} \cite{liu2014discrete}&supervised&0.338&12.60&0.134\\
            Li \textit{et al.} \cite{li2015depth}&supervised&0.279&10.27&0.102\\
            Liu \textit{et al.} \cite{liu2016learning}&supervised&0.287&14.09&0.122\\
            Roy \textit{et al.} \cite{roy2016monocular}&supervised&0.260&12.40&0.119\\
			MS-CRF \textit{et al.}\cite{xu2017multi}&supervised&0.198&8.56&-\\
			Atapour-A \textit{et al.}\cite{atapour2018real}&unsupervised&0.423&9.002&-\\
			Laina \textit{et al.} \cite{laina2016deeper}&supervised&0.201&7.038&0.079\\
			DORN(VGG)\cite{fu2018deep}&supervised&0.238&10.01&0.087\\
			DORN(ResNet)\cite{fu2018deep}&supervised&0.162&7.32&0.067\\
			Hu et al. \cite{hu2019revisiting}&supervised&0.179&6.613&0.070\\
			Ours-GAN&supervised&0.158&6.139&0.067\\
			Ours-GAN&semi-supervised&\textbf{0.153}&\textbf{6.054}&\textbf{0.066}\\
			\hline
	\end{tabular}}
\end{table}
\begin{figure}
	\centering\includegraphics[width=3.5in]{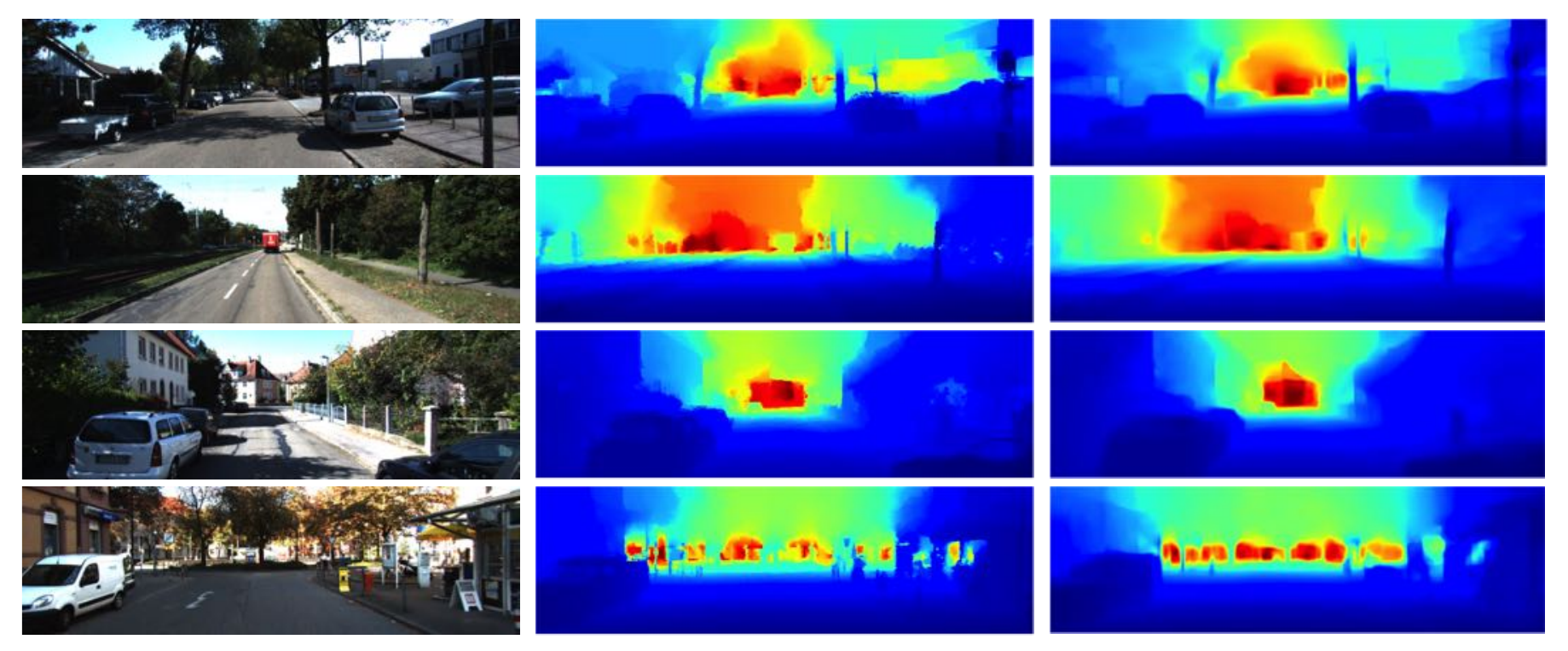}
	\caption{Qualitative results on KITTI dataset.
    The columns(from left to right) are RGB images, ground truth depth maps, and results by our semi-supervised model, respectively. Pixels with distance larger than $80m$ are masked out.}
	\label{fig:kitti}
\end{figure}

\subsection{KITTI Dataset}
\label{subsec:da}
In this section, we first evaluate our model in a larger data regime, using 22K image-depth pairs from the KITTI dataset \cite{Geiger2013IJRR} together with 20K additional RGB images from the Cityscape dataset \cite{Cordts2016Cityscapes} in our case. Evaluation results are reported in Tab. \ref{tab:kitti_quantitative}. Then, we experiment in a setting where domain shift (KITTI $\rightarrow$ Make3D in our case) exists. That is, we train on the KITTI image-depth pairs while test on the Make3D dataset.  
\begin{table}
	\caption{Comparison on the KITTI dataset. The values of the baselines are taken from the original papers.}
	\label{tab:kitti_quantitative}
	\scalebox{0.8}[0.8]{
		\begin{tabular}{p{2.7cm} p{3.6cm} p{0.8cm} p{0.8cm} p{1.0cm}}
			\hline
			algorithm&training data&rel&rms&rms $log$\\
			\hline
		    Eigen \textit{et al.}\cite{eigen2014depth}&KITTI&0.190&7.156&0.270\\
		    Zhou \textit{et al.}\cite{zhou2017unsupervised}&KITTI+Cityscape&0.198&6.565&0.275\\
		    Kuznietsov \textit{et al.}\cite{kuznietsov2017semi}&KITTI&0.108&\textbf{3.518}&0.176\\
		    Kumar \textit{et al.}\cite{cs2018monocular}&KITTI&0.219&6.340&0.273\\
		    Atapour-A \textit{et al.}\cite{atapour2018real}&KITTI+Cityscape&0.110&4.726&0.194\\
		    Ours-GAN&KITTI&0.107&4.405&0.181\\
		    Ours-GAN&KITTI+Cityscape (20k RGBs)&\textbf{0.093}&4.195&\textbf{0.170}\\
			\hline
	\end{tabular}}
\end{table}

The KITTI \cite{Geiger2013IJRR} dataset is captured by two high-resolution color (and grayscale) cameras and a Velodyne laser scanner on a driving car.
The raw dataset contains $61$ scenes categorized as \emph{City}, \emph{Residential}, \emph{Road} or \emph{Campus}.
Following \cite{eigen2014depth}, we use $33$ scenes of the dataset for training and leave the rest $28$ scenes for testing, which results in $22$K image-depth pairs in total.
The raw dataset stores depth by saving 3D points that are sampled by a rotating LIDAR scanner.
The ground truth depth maps are generated by reprojecting these points to the left RGB images using the given intrinsics and extrinsics. All image data and Velodyne depth measures have a spatial resolution of $1,242 \times 375$. Besides the training image-depth pairs drawn from the KITTI dataset, we also leverage more RGB images from the Cityscape dataset \cite{Cordts2016Cityscapes}. The Cityscape dataset offers dense pixel annotations of urban street scenes that are similar to the KITTI dataset. We draw 20k RGB images from the dataset and combine them with the 22k KITTI image-depth pairs to train our model in the semi-supervised manner.  A validation set of 160 samples drawn from the KITTI dataset is used to tune hyper-parameters. We test our model on the commonly-used Eigen split \cite{eigen2014depth}, which contains $697$ images selected from $28$ scenes.
During testing, we mask out the pixels with distance $\leq 0$ or $\geq 80$ meters from the ground truth depth map, and do not consider them in the subsequent error computation.
We report the quantitative comparisons with the state-of-the-art methods in Tab.~\ref{tab:kitti_quantitative}. By leveraging extra RGB images through the semi-supervised learning, our model can further decrease the evaluation error down to a new level. We also give some qualitative results in Fig. \ref{fig:kitti}. It is shown that predicted maps are sometimes even more natural than ground truth depth maps, due to the fact that ground truth depths given by KITTI are sparse and need to be densified by in-painting for visualization, while our model directly predicts dense depth values.

\begin{table}
	\caption{Adaptation results from the KITTI to the Make3D. Unlike other supervised methods, our model is capable to leverage the images (without their corresponding depth maps) in Make3D and adapts best to it.}
	\label{tab:make3d_adaptation}
	\scalebox{0.8}[0.8]{
		\begin{tabular}{p{1.9cm} p{3.8cm} p{1.0cm} p{1.0cm} p{1.0cm}}
			\hline
			algorithm&training data&rel&rms&$log_{10}$\\
			\hline
			Laina \textit{et al.} \cite{laina2016deeper}&KITTI&0.587&17.957&\textbf{0.223}\\
			DORN \cite{fu2018deep}&KITTI&0.589&10.701&0.239\\
			Ours&KITTI&0.555&12.153&0.387\\
			Ours&KITTI+Make3D (400 RGBs)&\textbf{0.447}&\textbf{10.349}&0.236\\
			\hline
	\end{tabular}}
\end{table}

We also evaluate our model in the viewpoint of domain adaptation. Domain adaptation is necessary when the training data distribution differs from the distribution of the testing data, which may cause significant performance drop during the algorithm deployment, as is shown in Tab. \ref{tab:make3d_adaptation}. The problem can be partially solved by our proposed semi-supervised learning in which RGB images drawn from unseen scenes are leveraged. We train on the image-depth pairs from KITTI along with images from Make3D, and then test the learned model on Make3D. To make the predicted maps of Make3D images comparable with those of KITTI images, we randomly crop a region on KITTI images in training, whose spatial size is aligned with the image size of Make3D. As shown in Tab.~\ref{tab:make3d_adaptation}, our model outperforms the models that are trained on KITTI in a supervised manner and directly tested on Make3D. 

\begin{figure}
	\centering\includegraphics[width=3in]{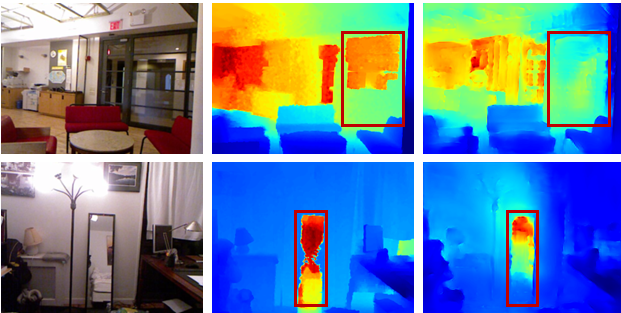}
	\caption{Failures in glass and mirror area (from left to right: RGB image, ground truth depth and predicted depth).}
	\label{fig:limitation}
\end{figure}
\subsection{Model Limitations}
\label{subsec:fails_and_limits}
We further take a deeper look into the cases when our model is less effective:

\textbf{Abundant Training GTs are available.} As shown in Fig. \ref{fig:nyud_comp}, when the model is fed with sufficient data, adding more unlabeled image shows no help to the prediction accuracy.
We experiment with three models in Tab.~\ref{tab:model_improvement} and find that our semi-supervised framework helps to improve the prediction accuracy in the case where data number is still the first driving force for model training.
After a saturation point of the training data, giving more unlabeled images does not help the model to learn better.
We argue that, this is expected for a semi-supervised learning approach, which has assumptions that the annotations are not sufficient.
When the training data is abundant, such assumptions are violated, and thus we cannot expect the semi-supervised approach can further outperform fully-supervised approaches.

\textbf{Predicting depth in translucency or high reflective regions.}
As illustrated in Fig. \ref{fig:limitation}, our model fails to predict the correct depth value in glass and mirror regions, due to the flaw of the training data. IR based depth acquisition equipment, such as Kinect, may collect misleading depth values in the unreflective region such as glass, and region with specular reflection such as mirror. With these training data flaws, the model can neither handle the depth prediction task in such regions.

\section{Conclusion}
\label{sec:con}
In this work, we propose an adversarial framework that produces state-of-the-art depth estimation results by using only a limited amount of training samples.
We achieve this by passing the feedbacks of two different discriminators (one inspects image-depth pairs and the other inspects depth channel only) to the generator as a unified loss, which makes effective use of unlabeled RGB images.
The experimental results on NYU Depth, Make3D and KITTI datasets demonstrate that our method with the proposed GAN loss can achieve better quantitative results and better qualitative results.
Moreover, our method generalizes well to both indoor and outdoor scenes. 
Our model also has less demand for labeled data and has the simplicity of adapting to various depth estimation models as the generator, which is therefore more flexible and robust in real-world applications.

\ifCLASSOPTIONcompsoc
  \section*{Acknowledgments}
\else
  \section*{Acknowledgment}
\fi
This work is supported by the National Key R\&D Program (No.2017YFC0113000, and No.2016YFB1001503), Nature Science Foundation of China (No.U1705262, No.61772443, and No.61572410).


\ifCLASSOPTIONcaptionsoff
  \newpage
\fi



%
%
%

\bibliographystyle{IEEEtran}
\bibliography{mybib}

\end{document}